\documentclass[final,12pt]{colt2022} 

\usepackage{bm}

\newcommand{\E}{\mathbb{E}}
\newcommand{\Prob}{\mathbb{P}}

\title[Optimal Best Arm Identification with Fixed Budget]{Open Problem: Optimal Best Arm Identification with Fixed Budget}
\usepackage{times}

\coltauthor{%
 \Name{Chao Qin} \Email{cq2199@columbia.edu}\\
 \addr Columbia University
}

\begin{document}

\maketitle

\begin{abstract}%
\emph{Best arm identification} or \emph{pure exploration} problems have received much attention in the COLT community since \citet{bubeck2009pure} and \citet{DBLP:conf/colt/AudibertBM10}. For any bandit instance with a unique best arm, its asymptotic complexity in the so-called \emph{fixed-confidence setting} has been completely characterized in \citet{GarivierK16} and \citet{chernoff1959sequential}, while little is known about the asymptotic complexity in its ``dual" setting called \emph{fixed-budget setting}.
This note discusses the open problems and conjectures about the instance-dependent asymptotic complexity in the fixed-budget setting.
\end{abstract}


\section{Introduction and problem formulation}
We consider the so-called best arm identification (BAI) or pure exploration problems where there is a finite number of arms.
An experimenter can sequentially select arms to measure and observes independent noisy observations of their quality. The experimenter's goal is to confidently identify a best arm through allocating measurement effort in an adaptive and intelligent manner. 
BAI problems have also been studied under different names for several decades, e.g., \emph{ranking and selection} or \emph{ordinal optimization} in the literature of statistics and operations research. 
The literature of machine learning mainly studies BAI problems in two settings. One is called \emph{fixed-confidence setting} where the objective is minimizing the expected number of collected samples while guaranteeing the probability of incorrect decision after the stopping time less than a pre-specified level, and the other is called \emph{fixed-budget setting} where the objective is minimizing the probability of incorrect decision after a given budget of samples is used up. 
For any bandit instance with a unique best arm, its asymptotic complexity in the fixed-confidence setting has been fully characterized.
See for example, \citet{GarivierK16} and \citet{chernoff1959sequential}. Although both settings seem ``dual" to each other, the instance-dependent asymptotic complexity in the fixed-budget setting is unclear for a very long time.
This note briefly include the existing results in the fixed-confidence setting and discusses the open problems and conjectures about the instance-dependent asymptotic complexity in the fixed-budget setting.

We use bold letters to denote vectors. A bandit instance $\bm{\mu}$ consists of $k$ unknown distributions or arms $\bm{\mu}=(\mu_1,\ldots,\mu_k)$ with respective expectations $\bm{\theta}=(\theta_1,\ldots,\theta_k)$. For the ease of exposition, we assume the bandit instance $\bm{\mu}$ has a unique best arm. Denote it by $I^*(\bm{\mu}) \triangleq \arg\max_{i\in[k]}\theta_i$ where $[k] \triangleq\{1,\ldots,k\}$. 
The bandit instance $\bm{\mu}$ is \emph{unknown} to an experimenter who wants to confidently identify the best arm $I^*(\bm{\mu})$ at the end of the experiment. At each time $t=1,2,\ldots$, according to the information collected so far, she can choose an arm $I_t\in [k]$ to measure and then observes an independent noisy observation $Y_{t,I_t}$ drawn from distribution $\mu_{I_t}$.

\section{Fixed-confidence setting and its known results} 

In the fixed-confidence setting, the experimenter can stop gathering samples at any time and returns an estimate of the identity of the best arm after that. The experimenter's algorithm is then composed of three rules:
a sampling rule that determines which arm to sample at each time, a stopping rule that decides whether to stop at each time,
and a decision rule that at the stopping time $\tau$, returns an estimate $\hat{I}_{\tau}$ of the identity of the best arm based on the first $\tau$ observations.

Let $\mathcal{S}$ be the class of bandit instances with a unique best arm. \citet{GarivierK16} studies algorithms that guarantee a \emph{uniformly} small probability of incorrect decision (at the stopping time) below a pre-specified level $\delta>0$, in the sense that 
\begin{equation}
\label{eq:delta PAC}
\forall \bm{\mu}\in\mathcal{S}, \quad \Prob_{\bm{\mu}}\left(\hat{I}_{\tau_\delta} \neq I^*(\bm{\mu})\right)\leq \delta
\end{equation}
where $\tau_\delta$ is an \emph{almost surely finite} stopping time. 
The notation $\Prob_{\bm{\mu}}(\cdot)$ indicates that we are evaluating the probability of events when the observations from chosen arms are drawn under the bandit instance $\bm{\mu}$.
In the learning theory literature, such algorithms are called \emph{$\delta$-Probably-Approximately-Correct} or \emph{$\delta$-PAC}. Among such algorithms, we would like to minimize the expected number of collected samples denoted by $\E_{\bm{\mu}}[\tau_\delta]$.  \citet{GarivierK16} shows that for any $\delta$-PAC algorithm,
\begin{equation}
\label{eq:fixed confidence lower bound}
\forall \bm{\mu}\in\mathcal{S}, \quad \liminf_{\delta\to 0}  \frac{\E_{\bm{\mu}}[\tau_\delta]}{\log(1/\delta)} \geq \Gamma^*_\mathrm{fc}({\bm{\mu}})
\end{equation}
where 
\begin{equation}
\label{eq:fixed confidence complexity}
\Gamma^*_\mathrm{fc}({\bm{\mu}}) = \left( \sup_{\bm{w}\in\Sigma_k} \inf_{ \bm{\nu}\in \text{Alt}(\bm{\mu})}  \sum_{i=1}^k  w_i\mathrm{KL}(\mu_i \|\nu_i)   \right)^{-1}.
\end{equation}
Here $\Sigma_k$ is the probability simplex of dimension $k-1$; $\text{Alt}(\bm{\mu})\triangleq \{\bm{\nu}\in\mathcal{S}: I^*(\bm{\nu})\neq I^*(\bm{\mu})\}$ is the set of bandit instances whose unique best arm is different from $\bm{\mu}$'s unique best arm; $\mathrm{KL}(p \| q)$ denotes the Kullback-Leibler (KL) divergence between distributions $p$ and $q$. The subscript $\mathrm{fc}$ in  $\Gamma^*_\mathrm{fc}$ is the acronym of ``fixed-confidence".
Besides the information-theoretic lower bound in Equation \eqref{eq:fixed confidence lower bound}, \citet{GarivierK16} also proposes the so-called Track-and-Stop algorithms that are $\delta$-PAC and can guarantee
\begin{equation}
\label{eq:fixed confidence upper bound}
\forall \bm{\mu}\in\mathcal{S}, \quad \limsup_{\delta\to 0}  \frac{\E_{\bm{\mu}}[\tau_\delta]}{\log(1/\delta)} \leq \Gamma^*_{\mathrm{fc}}({\bm{\mu}}).
\end{equation}
Since the lower and upper bounds in Equations \eqref{eq:fixed confidence lower bound} and \eqref{eq:fixed confidence upper bound} are the same, the function $\Gamma^*_\mathrm{fc}: \mathcal{S}\to\mathbb{R}$ characterizes the asymptotic complexity in the fixed-confidence setting.

\section{Fixed-budget setting and its open problems}

In the fixed-budget setting, a budget of $n$ samples is fixed and given. 
After collecting $n$ samples, the experimenter needs to decide an estimate of the identity of the best arm denoted by $\hat{I}_{n}$. An algorithm is then only consists of a sampling rule and a decision rule. The experimenter's objective in the fixed-budget setting is to minimize the probability of incorrect decision defined as
\[
p_{\bm{\mu}, n} \triangleq \Prob_{\bm{\mu}}\left(\hat{I}_{n}\neq I^*(\bm{\mu})\right).
\]
This setting seems ``dual" to the fixed-confidence setting in the sense that instead of minimizing the number of samples subject to a uniformly small probability of incorrect decision, here we minimize the probability of incorrect decision subject to a fixed budget of samples. However, little is known about the asymptotic complexity in the fixed-budget setting.

\paragraph{Open problem 1.} The first and foremost open problem is whether there are a desirable algorithm class $\mathcal{A}$ and a well-defined function $\Gamma^*_\mathrm{fb}: \mathcal{S}\to\mathbb{R}$ such that for any algorithm in $\mathcal{A}$,
    \[
    \forall \bm{\mu}\in\mathcal{S},\quad
    \liminf_{n\to\infty}  \frac{n}{\log(1/p_{\bm{\mu},n})}\geq \Gamma^*_\mathrm{fb}({\bm{\mu}})
    \]
and there is an algorithm that belongs to $\mathcal{A}$ and guarantees
    \[
    \forall \bm{\mu}\in\mathcal{S},\quad
    \limsup_{n\to\infty}  \frac{n}{\log(1/p_{\bm{\mu},n})}\leq \Gamma^*_\mathrm{fb}({\bm{\mu}}).
    \]
Here the subscript $\mathrm{fb}$ in $\Gamma^*_\mathrm{fb}$ is the acronym of ``fixed-budget".    

\paragraph{Discussion on potential algorithm class.} \citet{JMLR:v17:kaufman16a} studies the so-called \emph{consistent} algorithms such that
for any $\bm{\mu}\in\mathcal{S}$, the probability of incorrect decision $p_{\bm{\mu},n}$ goes to zero when $n$ increases to infinity. 
The class of consistent algorithms is relatively large, and we believe it might not be the right algorithm class for characterizing the asymptotic complexity in the fixed-budget setting.
Note that in the fixed-confidence setting, the class of $\delta$-PAC algorithms defined in Equation \eqref{eq:delta PAC} is restrictive in the sense that it requires a uniformly small probability of incorrect decision (at the stopping time) for any bandit instance $\bm{\mu}\in\mathcal{S}$.  
This restriction helps the analysis of the asymptotic complexity in the fixed-confidence setting.
We believe it is necessary to come up with a natural but more restrictive algorithm class in the fixed-budget setting. For example, besides the convergence of the probability of incorrect decision to zero, we may also need to control the convergence rate of the algorithms in the class. 
One potential algorithm class contains all the algorithms that perform uniformly no worse than uniform sampling, i.e., for any algorithm in this class, it achieves a lower or the same value of $\limsup_{n\to\infty}  \frac{n}{\log(1/p_{\bm{\mu},n})}$ for any bandit instance $\bm{\mu}\in\mathcal{S}$. This leads to the following open problem.

\paragraph{Open problem 2.} This open problem is whether there is an algorithm other than uniform sampling itself that performs uniformly no worse than uniform sampling in the fixed-budget setting.

Indeed in the fixed-confidence setting, one can show that for any bandit instance $\bm{\mu}$, $\Gamma^*_\mathrm{fc}({\bm{\mu}})$ is  less than or equal to the value of $ \limsup_{n\to\infty}\frac{\E_{\bm{\mu}}[\tau_\delta]}{\log(1/\delta)}$ under uniform sampling. This implies those asymptotically optimal algorithms in the fixed-confidence setting perform uniformly no worse than uniform sampling. We tend to believe that those algorithms also have advantages over uniform sampling in the fixed-budget setting, but the answer to this open problem is unclear.

\section{Conjectures}
In this section, we state two existing conjectures in the literature. Unfortunately, neither of them is correct in general.

\paragraph{Conjecture 1.}
Since the fixed-budget and fixed-confidence settings are ``dual" to each other, one conjecture is that $\Gamma^*_\mathrm{fb} =\Gamma^*_\mathrm{fc}$.

\paragraph{Conjecture 2.}
Another conjecture is that  $\Gamma^*_\mathrm{fb} =\Gamma^*_\mathrm{na}$ where $\Gamma^*_\mathrm{na}$ defined later is the asymptotic complexity in a \emph{non-adaptive} version of the fixed-budget setting studied in \citet{glynn2004large} (and the subscript $\mathrm{na}$ in $\Gamma^*_\mathrm{na}$ is the acronym of ``non-adaptive").
They consider sampling rules that fix the probability vector $\bm{w}$ of selecting $k$ arms in each time and thus do not adapt to the observations from sequentially selected arms. 
They show that
for any bandit instance $\bm{\mu}\in\mathcal{S}$, 
    \[
    \forall\bm{w}\in\Sigma_k, \quad
    \liminf_{n\to\infty}  \frac{n}{\log(1/p_{\bm{\mu},n})}\geq \Gamma^*_\mathrm{na}({\bm{\mu}})
    \]
    and
    \[
    \exists \bm{w}^*(\bm{\mu})\in\Sigma_k, \quad
    \limsup_{n\to\infty}  \frac{n}{\log(1/p_{\bm{\mu},n})}\leq \Gamma^*_\mathrm{na}({\bm{\mu}})
    \]
    where 
    \begin{equation}
    \label{eq:non adaptive complexity}
    \Gamma^*_\mathrm{na}({\bm{\mu}}) = \left( \sup_{\bm{w}\in\Sigma_k} \inf_{ \bm{\nu}\in \text{Alt}(\bm{\mu})}  \sum_{i=1}^k  w_i\mathrm{KL}( \nu_i \| \mu_i)   \right)^{-1}.
    \end{equation}
At the first glance, the complexity term $\Gamma^*_\mathrm{na}({\bm{\mu}})$ in Equation \eqref{eq:non adaptive complexity} looks the same as $\Gamma^*_\mathrm{fc}({\bm{\mu}})$ in Equation \eqref{eq:fixed confidence complexity}. Indeed they are different since KL divergence is not symmetrical in general, 
but for Gaussian distributions, $\Gamma^*_\mathrm{na}({\bm{\mu}}) = \Gamma^*_\mathrm{fc}({\bm{\mu}})$.
Note that the optimal sampling vector $\bm{w}^*(\bm{\mu})$ depends on the knowledge of unknown bandit instance $\bm{\mu}$, so it is unknown a priori. Hence, the sampling rule that always fixes the optimal sampling vector $\bm{w}^*(\bm{\mu})$ for each bandit instance $\bm{\mu}$ is not a valid choice for the adaptive fixed-budget setting of our interest.

\paragraph{Neither conjecture is correct.}
The results in \citet{https://doi.org/10.48550/arxiv.2109.08229} imply neither conjecture is correct in general for Bernoulli bandits. Inspired by the construction in \citet{pmlr-v49-carpentier16}, \citet{https://doi.org/10.48550/arxiv.2109.08229} constructs a set of bandit instances with large number of arms and shows that neither conjecture can hold for all the instances.
We believe that one can also show similar negative results for Gaussian bandits.

\section{Known results for two-armed bandits}

Though neither conjecture is correct in general, \citet{JMLR:v17:kaufman16a} shows that both conjectures hold for two-armed Gaussian bandits with known variances, i.e., $\Gamma^*_\mathrm{fb}(\bm{\mu}) =\Gamma^*_\mathrm{fc}(\bm{\mu}) = \Gamma^*_\mathrm{na}(\bm{\mu})$ for any such bandit instance $\bm{\mu}$.
It further proves that the optimal sampling rule is non-adaptive, which fixes the sampling vector $(\frac{\sigma_1}{\sigma_1 + \sigma_2}, \frac{\sigma_2}{\sigma_1 + \sigma_2})$ where $\sigma_1$ and $\sigma_2$ are the known variances of the two arms. Recently, \citet{Kato2022small} shows that when the gap between the unknown means of the two arms goes to zero, even the variances are also unknown, the upper bound of the proposed algorithm matches the instance-dependent lower bound in \citet{JMLR:v17:kaufman16a}.  \citet{adusumilli2022minimax} studies the diffusion regime of two-armed Gaussian bandits and proves that the same sampling vector $(\frac{\sigma_1}{\sigma_1 + \sigma_2}, \frac{\sigma_2}{\sigma_1 + \sigma_2})$ is also minimax optimal.
However, for two-armed Bernoulli bandits, \citet{JMLR:v17:kaufman16a} shows that though the optimal sampling vector exists, it requires the knowledge of unknown means of the arms, which is unknown a priori. It is unclear whether there is an algorithm can achieve the asymptotic optimality without such a requirement.

\acks{We thank Kaito Ariu, R{\'e}my Degenne, Sandeep Juneja,  Masahiro Kato, Junpei Komiyama, Wouter M. Koolen, Pierre M{\'e}nard, Daniel Russo and Assaf Zeevi for fruitful discussions.}

\bibliography{references}

\end{document}